Topic: Foundations of Probability Theory for A.I.

Title: THE APPLICATION OF ALGORITHMIC PROBABILITY TO PROBLEMS IN ARTIFICIAL INTELLIGENCE*

Author: Ray Solomonoff, Oxbridge Research Box 559, Cambridge, Mass. 02238

INTRODUCTION: This paper covers two topics: first an introduction to Algorithmic Complexity Theory: how it defines probability, some of its characteristic properties and past successful applications.

Second, we apply it to problems in A.I. - where it promises to give near optimum search procedures for two very broad classes of problems.

Algorithmic probability of a string, s, is the probability of that particular string being produced as the output of a reference universal Turing machine with random input. It is approximately $2^{-\ell(p)}$, where $\ell(p)$ is the length of the shortest program, p, that will produce s as output. $\ell(p)$ is the Kolmogorov complexity of s — one form of Algorithmic complexity.

Algorithmic complexity has been applied to many areas of science and mathematics. It has given us a very good understanding of randomness [6]. It has been used to generalize information theory [2] and has clarified important concepts in the foundations of mathematics [1]. It has given us our first truly complete definition of probability [7,8,11].

The completeness property of Algorithmic probability means that if there is any describable regularity in a body of data, our system is guaranteed to discover it using a relatively small sample of the data. It is the only probability evaluation method known to be complete. As a necessary consequence of its completeness, this kind of probability must be incomputable. Conversely, any computable probability measure cannot be complete.

Can we use this incomputable probability measure to obtain solutions to practical problems? A large step in this direction was Levin's search procedure that obtains a solution to any P or NP problem within a constant factor of optimum time. The "constant factor" may, in general, be quite large. While this technique does not use a complete probability measure, it uses a measure that approaches completeness.

Under certain reasonable conditions and for a broader class of problems than Levin originally considered, the "constant factor" must be less than about four.

The P or NP class originally considered contained machine inversion problems: we are given a string, s, and a machine, M, that maps strings to strings. We must find in minimum time, a string, x, such that $M(x) = s$. Solving algebraic equations, symbolic integration and theorem proving are examples of this broad class of problems.

However, Levin's search procedure also applies to another broad class of problems — Time limited optimization problems. Given a time limit T, and a machine M that maps strings to real numbers, to find within time T the string x such that M(x) is as large as possible. Many engineering problems are of this sort - for example, designing an automobile in 6 months satisfying certain specifications having minimum cost. Constructing the best possible probability distribution or physical theory from empirical data in limited time is also of this form.

In solving either machine inversion problems or time limited optimization problems, it is usually possible to express the information needed to solve the problem (either heuristic information or problem-specific information) by means of a conditional probability distribution. This distribution relates the string that describes the problem to any string that is a candidate solution to the problem. If all of the information needed to solve the problem is in this probability distribution and we do not modify the distribution during the search, then Levin's search technique is within a factor of 2 of optimum.

If we are allowed to modify the probability distribution during the search on the basis of our experience in attempting to solve the problem, then Levin's technique is within a factor of 4 of optimum.

The efficacy of this problem solving technique hinges on our ability to represent all of our relevant knowledge in a probability distribution.

To what extent is this possible? For one broad area of knowledge this is certainly easy to do: this is the kind of inductive knowledge obtained from a set of examples of correctly worked problems. Algorithmic probability obtained from examples of this sort is in just the right form for application of our general problem solving system. Furthermore, when we have other kinds of information that we want to express as a probability distribution we can usually hypothesize a sequence of examples that would lead to the learning of that information by a human. We can then give that set of examples to our induction system and it will acquire the same information in appropriate probabilistic form.

While it is possible to put most kinds of information into probabilistic form using this technique, a person can, with some experience, learn to bypass this process and express the desired information directly in probabilistic form. We will show how this can be done for certain kinds of heuristic information such as Planning, Analogy, Clustering and Frame Theory.

The use of a probability distribution to represent knowledge, not only simplifies the solution of problems, but it enables us to put information from many different kinds of problem solving systems into a common format. Then, using techniques that are fundamental to algorithmic complexity theory, we can compress this heterogeneous mass of information into a more compact, unified form. This operation corresponds to Kepler's laws summarizing and compressing Tycho Brahe's empirical data on planetary motion. Algorithmic complexity theory has this ability to synthesize, to find general laws in masses of unorganized and partially organized knowledge. It is in this area that its greatest value for A.I. lies.

I will conclude with a discussion of the present state of the system and the outstanding problems.

---

* Much of the content of sections I and II was presented at a workshop on "Theories of Complexity", Cambridge, Mass., August, 1984.

# I. ALGORITHMIC COMPLEXITY

The earliest application of algorithmic complexity was to devise a formal theory of inductive inference [7,8,11]. All induction problems are equivalent to the problem of extrapolating a long sequence of symbols. Formally, we can do this extrapolation by Baye's theorem, if we are able to assign an apriori probability to any conceivable string of symbols, x.

This can be done in the following manner: Let x be a string of n binary symbols. Let M be a universal Turing machine with 3 tapes: a unidirectional input tape; unidirectional output tape; and an infinitely long bidirectional work tape.

The unidirectionality of the input and output tapes assure us that if $M(s) = y$, then $M(ss') = yy'$ — i.e. if s is the code of y, then if we extend s by several symbols, the output of M will be at least y and may possibly (though not necessarily) be followed by other symbols.

We assign an apriori probability to the string x, by repeatedly flipping a coin and giving the machine M, an input 1 whenever we have "heads" or 0 whenever we have "tails". There is some probability $P_M(x)$ that this random binary input string will cause M to have as output a string whose first n bits are identical to x. When constructed in this manner with respect to universal Turing machine, M, $P_M(x)$ becomes the celebrated universal apriori probability distribution.

Conceptually, it is easy to calculate $P_M(x)$. Suppose $s_1, s_2, s_3 \ldots$ are all of the possible input strings to M that can produce x (at least) as output. Let $s_1', s_2', s_3' \ldots$ be a maximal subset of $\{s_i\}$ such that no $s_i'$ can be formed by adding bits onto the end of some other $s_j'$. Thus no $s_i'$ can be the "prefix" of any other $s_j'$.

The probability of $s_i'$ being produced by random coin tossing is just $2^{-\ell(s_i')}$, where $\ell(s_i')$ is the number of bits in $s_i'$.

Because of the prefix property, the $s_i$ are mutually exclusive events and so the probability of x being produced by any of them is simply

$$\sum_i 2^{-\ell(s_i)}$$

which is therefore the value of $P_M(x)$.

To do prediction with $P_M(x)$ is very simple.

The probability that x will be followed by 1 rather than 0 is

$$P_M(x1)/(P_M(x0) + P_M(x1))$$

How accurate are these probabilities? Suppose that $P(a_{n+1} = 1 | a_1 a_2 a_3 \ldots a_n)$ is a conditional probability distribution for the $n+1^{th}$ bit of a binary string, given the previous n bits, $a_1 a_2 a_3 \ldots a_n$. Let us further postulate that P is describable by machine M with a program b bits long.

Let $P_M(a_{n+1} = 1 | a_1 a_2 \ldots a_n)$ be the corresponding probability distribution based on $P_M$. Using P, and a suitable source of randomness, we can generate a stochastic sequence $A = a_1 a_2 a_3 \ldots a_m$. Both P and $P_M$ are able to assign probabilities to the occurrence of the symbol 1 at any point in the sequence A based on the previous symbols in A.

It has been shown [11, pp. 426-427] that the total expected squared error between P and $P_M$ is given by

$$E_P \sum_{m=1}^{n} (P_M(a_{m+1} = 1 | a_1 a_2 \ldots a_m) - P(a_{m+1} = 1 | a_1 a_2 \ldots a_m))^2 < b \ln \sqrt{2}$$

The expected value is with respect to probability distribution, P.

This means that the expected value of the sum of the squares of the deviations of $P_M$ from P is bounded by a constant.

This error is much less than that given by conventional statistics — which is proportional to ln n. The disparity is because P is describable by a finite string of symbols. Usually statistical models have parameters that have an infinite number of bits in them and so the present analysis must be applied to them in somewhat modified form. The smallness of this error assures us that if we are given a stochastic sequence created by an unknown generator, we can use $P_M$ to obtain the conditional probabilities of that generator with much accuracy.

Tom Cover [3, also 11, pp. 425] has shown that if $P_M$ is made the basis of a universal gambling scheme, its yield will be extremely large.

It is clear that $P_M$ depends on just what universal machine M, is used. However, if we use a lot of data for our induction, then the probability values are relatively insensitive to choice of M. This will be true even if we include as data, information not directly related to the probabilities we are calculating.

We believe that $P_M$ gives us about the best probabiliity values that are obtainable with the available information.

While $P_M$ has many desireable properties, it cannot ever be used directly to obtain probability values. As a necessary consequence of its "completeness" — its ability to discover the regularities in any reasonable sample of data — $P_M$ must be uncomputable. However, approximations to $P_M$ are always possible and we will later show how to obtain close to the best possible approximations with given computational resources.

One common way to obtain approximations of a probability distribution to extrapolate the string, x, is to obtain short codes for x. In general, short programs for the sequence x correspond to regularities in x. If x was a sequence of a million 1's we could describe x in a few words and write a short program to generate it. If x was a random sequence with no regularities, then the shortest description of x would be x itself. Unfortunately, we can never know that a sequence is random. All we can ever know is that we have spent a lot of time looking for regularities in it and we've not found any. However, no matter how long we have looked, we can't be sure that we wouldn't find a regularity if we looked for 10 minutes more!

Any legitimate regularity in x can be used to write a shorter code for it. This makes it possible to give a clear criterion for success to a machine that is searching for regularities in a body of data. It is an adequate basis for the mechanization of inductive inference.

## II. A GENERAL SYSTEM FOR SOLVING PROBLEMS

The problems solvable by the system fall in two broad classes: machine inversion problems and time limited optimization problems. In both, the problem itself as well as the solution, can be represented by a finite string of symbols.

We will try to show that most, if not all knowledge needed for problem solving can be expressed as a conditional probability distribution relating the problem string (condition) to the probability of various other strings being solutions. We shall be interested in probability distributions that list possible solutions with their associated probability values in decreasing order of probability.

We will use Algorithmic complexity theory to create a probability distribution of this sort.

Then, considerations of Computational Complexity lead to a near optimum method to search for solutions.

We will discuss the advantages of this method of knowledge representation - how it leads to a method of unifying the Babel of disparate techniques used in various existing problem solving systems.

### Kinds of Problems that the System Can Solve.

Almost all problems in science and mathematics can be well approximated or expressed exactly as either machine inversion problems or time limited optimization problems.

Machine inversion problems include NP and P problems. They are problems of finding a number or other string of symbols satisfying certain specified constraints. For example, to solve $x + \sin x = 3$, we must find a string of symbols, i.e. a number, x that satisfies this equation.

Problems of this sort can always be expressed in the form $M(x) = c$. Here M is a computing machine with a known program that operates on the number x. The problem is to find an x such that the output of the program is c.

Symbolic integration is another example of machine inversion. For example we might want the indefinite integral of $xe^{x^2}$. Suppose M is a computer program that operates on a string of symbols that represent an algebraic expression and obtains a new string of symbols representing the derivative of the input string. We want a string of symbols, s such that $M(s) = xe^{x^2}$.

Finding proofs of theorems is also an inversion problem.

Let Th be a string of symbols that represents a theorem.

Let Pr be a string of symbols that represents a possible proof of theorem Th.

Let M be a program that examines Th and Pr. If Pr is a legal proof of Th, then its output is "Yes", otherwise it is "No".

The problem of finding a proof becomes that of finding a string s such that
$M(Th,s) = Yes$.

There are very many other problems that can be expressed as machine inversion problems.

Another broad class of problems are time limited optimization problems. Suppose we have a known program M, that operates on a number or string of symbols and produces as output, a real number between zero and one. The problem is to find an input that gives the largest possible output and we are given a fixed time, $T$, in which to do this.

Many engineering problems are of this sort. For example, consider the problem of designing a rocketship satisfying certain specifications, having minimal cost, within the time limit of 5 years.

Another broad class of optimization problems of this sort are induction problems. An example is devising the best possible set of physical laws to explain a certain set of data and doing this within a certain restricted time.

It should be noted that devising a good functional form for M - the criterion for how good the theory fits the data, is not in itself part of the optimization problem. A good functional form can, however, be obtained from algorithmic complexity theory.

The problem of extrapolating time series involves optimizing the form of prediction function, and so it, too can be regarded as an optimization problem.

Another form of induction is "operator induction". Here we are given an unordered sequence of ordered pairs of objects such as (1,1), (7,49), (-3,9), etc. The problem is to find a simple functional form relating the first element of each pair (the "input") to the second element (the "output"). In the example given, the optimum is easy to find, but if the functional form is not simple and noise is added to the output, the problems can be quite difficult.

Some "analogy" problems on I.Q. tests are forms of operator induction.

In the most general kind of induction, the permissible form of the prediction function is very general, and it is impossible to know if any particular function is the best possible - only that it is the best found thus far. In such cases the unconstrained optimization problem is undefined and including a time limit constraint is one useful way to give an exact definition to the problem.

All of these constrained optimization problems are of the form: given a program M, to find string x in time $T$ such that $M(x)$ = maximum. In the examples given, we always knew what the program M was. However, in some cases M may be a "black box" and we are only allowed to make trial inputs and remember the resultant outputs. In other forms of the optimization problem, M may be time varying and/or have a randomly varying component.

We will discuss at this time only the case in which the nature of M is known and is constant in time. Our methods of solution are, however, applicable to certain of the other cases.

In both inversion and optimization problems, the problem itself is represented as a string of symbols. For inversion problem, $M(x) = c$, this will consist of the program M followed by the string, c.

For optimization problems $M(x) = max$, in time $T$, our problem is represented by the program M followed by the number, $T$.

For inversion problems, the solution, x, will always be a string of the required form.

For an optimization problem, a solution will be a program that looks at the program M, and time $\tau$, and results of previous trials and from these creates as output, the next trial input to M. This program is always representable as a string, just as is the solution to an inversion problem.

Before telling how to solve these two broad categories of problems, I want to introduce a simple theorem in probability.

At a certain gambling house there is a set of possible bets available — all with the same big prize. The $i^{th}$ possible bet has probability $p_i$ of winning and it costs $d_i$ dollars to make the $i^{th}$ bet. All probabilities are independent and one can't make any particular bet more than once. The $p_i$ need not be normalized.

If all the $d_i$ are one dollar, the best bet is clearly the one of maximum $p_i$. If one doesn't win on that bet, try the one of next largest $p_i$, etc. This strategy gives the least number of expected bets before winning.

If the $d_i$ are not all the same, the best bet is that for which $p_i/d_i$ is maximum. This gives the greatest win probability per dollar.

Theorem I: If one continues to select subsequent bets on the basis of maximum $p_i/d_i$, the expected total money spent before winning will be minimal.

In another context, if the cost of each bet is not dollars, but time, $t_i$, then the betting criterion $p_i/t_i$ gives least expected time to win.

In order to use this theorem to solve a problem we would like to have the functional form of our conditional probability distribution suitably tailored to that problem. If it were an inversion problem, defined by M and c, we would like a function with M and c as inputs, that gave us, as output a sequence of candidate strings in order of decreasing $p_i/t_i$. Here $p_i$ is the probability that the candidate will solve the problem, and $t_i$ is the time it takes to generate and test that candidate. If we had such a function, and it contained all of the information we had about solving the problem, an optimum search for the solution would simply involve testing the candidates in the order given by this distribution.

Unfortunately we rarely have such a distribution available — but we can obtain something like it from which we can get good solutions.

One such form has input M and c as before, but as output it has a sequence of string, probability pairs $(a_1,p_1), (a_2,p_2)\ldots$ . $p_i$ is the probability that $a_i$ is a solution and the pairs are emitted in order of decreasing $p_i$.

When algorithmic complexity is used to generate probability distributions, these distributions have approximately this form. How can we use such a distribution to solve problems?

First we select a small time limit, $T_0$ and we do an exhaustive search of all candidate solution strings, $a_i$ such that $t_i/p_i < T_0$. Here $t_i$ is the time needed to generate and test $a_i$. If we find no solution, we double $T_0$ and go through the exhaustive testing again. The process of doubling $T_0$ and searching is repeated until a solution is found.

The entire process is approximately equivalent to testing the $a_i$'s in order of increasing $t_i/p_i$.

It's not difficult to prove Theorem II.

Theorem II: If a correct solution to the problem is assigned a probability $p_j$ by the distribution, and it takes time $t_j$ to generate and test that solution, then the algorithm described will take a total search time of less than $2 t_j/p_j$ to find that solution.

Theorem III: If all of the information we have to solve the problem is in the probability distribution, and the only information we have about the time needed to generate and test a candidate is by experiment, then this search method is within a factor of 2 of the fastest way to solve the problem.

In 1973 L. Levin [5,13] used an algorithm much like this one to solve the same kinds of problems, but he did not postulate that all of the information needed to solve the problems was in the equivalent of the probability distribution. Lacking this strong postulate, his conclusion was weaker — i.e. that the method was within a constant factor of optimum. Though it was clear that this factor could often be very large, he conjectured that under certain circumstances it would be small. Theorem III gives one condition under which the factor is 2.

In artificial intelligence research, problem solving techniques optimum within a factor of 2 are normally regarded as much more than adequate, so a superficial reading of Theorem III might regard it as a claim to have solved most of the problems of A.I.! This would be an inaccurate interpretation.

Theorem III postulates that we put all of the needed problem solving information, (both general heuristic information as well as problem specific information) in the probability distribution. To do this, we usually use the problem solving techniques that are used in other kinds of problem solving systems and translate them into modifications of the probability distribution. The process is analogous to the design of an expert system by translating the knowledge of a human expert into a set of rules. While I have developed some standard techniques for doing this the translation process is not always a simple routine. Often it gives the benefit of viewing a problem from a different perspective, yielding new, better understanding of it. Usually it is possible to simplify and improve problem solving techniques a great deal by adding probabilistic information.

The Overall Operation of the System

We start with a probability distribution in which we have placed all of the information needed to solve the problems. This includes both general heuristic as well as problem specific information.

We also have the standard search algorithm of Theorem II which has been described.

The first problem is then given to the system. It uses the search algorithm and probability distribution to solve the problem.

If the problem is not solvable in acceptable time, then the $t_j/p_j$ of the solution must be too large. $t_j$ can be reduced by using a faster computer, or dividing up the search space between several computers, or by using faster algorithms for various calculations. $p_j$ can be increased by assigning short codes (equivalent to high probabilities) to commonly used sequences of operations. Reference 8, pp. 232-240, tells how to do this.

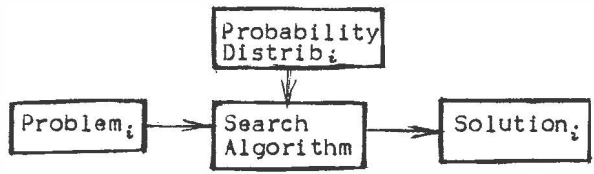

After the first problem is solved we have more information than we had before, (i.e. the solution to the first problem) and we want to put this information into the probability distribution. This done by a process of "compression". We start out with the probability distribution: It has been obtained by finding one or more short codes for a body of data, D . Let the single length equivalent of all of these codes be $L_0$ . Suppose the description length of our new problem, solution pair, PS , is $L_{ps}$ . "Compression" consists of finding a code for the compound object, (D, PS) that is of length less than $L_0 + L_{ps}$ . Compression is expressible as a time limited optimization problem. If the system took time T to solve the original problem, we will give it about time T for compressing the solution to this poblem into the probability distribution.

This compression process amounts to "updating" the probability distribution.

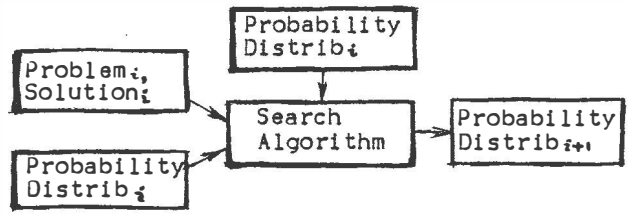

There may seem to be some logical difficulty in having a machine work on the improvement of a description of part of itself. However, we need not tell the machine that this is what it is doing. In the system described there is no way for it to obtain or use such information.

After compression, we give it the next external problem to solve - followed by another updating or compression session.

If the probability distribution contains little information relevant to compression, then the compression can be done by the system's (human) operator. Eventually the distribution will acquire enough information (through problem solving experience or through direct modification of the probability distribution by the operator) to be able to do useful compression in the available time without human guidance.

What are the principal advantages of expressing all of our knowledge in probabalistic form?

First: We have a near optimum method of solving problems in that form.

Second: It is usually not difficult to put our information into that form and when we do so, we often find that we can improve problem solving methods considerably.

Traditionally, a large fraction of A.I. workers have avoided probabilistic methods. By ignoring probability data they are simply throwing away relevant information. Using it in an optimum manner for search can do nothing less than speed up the search process and give a more complete understanding of it.

Third: Theorem II gives us a unique debugging tool. If the system cannot find a known solution to a problem in acceptable time, analysis of the $t_j$ and $p_j$ involved, will give us alternative ways to reduce $2 t_j/p_j$ .

Fourth: Once our information is in this common format, we can compress it. The process of compression of information involves finding shorter codes of that information. This is done by finding regularities in it and expressing these regularities as short codes.

Why do we want to compress the information in our probability distribution?

First: By compressing it, we find general laws in the data. These laws automatically interpolate, extrapolate and smooth the data.

Second: By expressing a lot of information as a much smaller collection of laws, we are in a far better position to find higher order laws than we would be if we worked with the data directly. Newton's laws were much easier to discover as an outgrowth of Kepler's laws, than it would be for Newton to derive them directly from purely experimental data.

Third: The compressed form of data is easier for humans to understand, so they may better know how to debug and improve the system.

Fourth: The processes of interpolating and extrapolating problem solving methods automatically creates new trial problem solving methods that have high probability of working. This makes it possible for the system to go beyond the insular techniques originally built into it and gives us the closest thing to true creativity that we can expect in a mechanized device.

In short, compression of information in the probability distribution transforms our system from a collection of disconnected problem solving techniques in to a unified, understandable system. It promises to make A.I. an integrated science rather than a compendium of individually developed, isolated methodologies.

III. USING PROBABILITY DISTRIBUTIONS TO REPRESENT KNOWLEDGE

A critical question in the forgoing discussion is whether we can represent all of the information we need to solve our problems through a suitable probability distribution. I will not try to prove this can always be done, but will give some fairly general examples of how to do it.

The first example will be part of the problem of learning algebraic notation from examples. The examples are of the form
```
    35,  41, + : 76
     8,   9, x : 72
    -8,   1, + : -7   etc.
```
The examples all use +, -, x, and ÷ only. The problem is for the machine to induce the relationship of the string to the right of the colon, to the rest of the expression.

To do this it has a vocabulary of 7 symbols:
$R_1$, $R_2$ and $R_3$, represent the first 3 symbols of its input.
Add, Sub, Mul, Div, represent internal operators that can operate on the contents of $R_1$, $R_2$, and $R_3$ if they are numbers.

The system tries to find a short sequence of these 7 symbols that represents a program expressing the symbol to the right of the colon in terms of the other symbols.

35, 41, + : 76 can be written as 35, 41, + : $R_1$, $R_2$, Add. If all symbols have equal probability to start, the subsequence
$R_1$, $R_2$, Add has probability
$1/7 \times 1/7 \times 1/7 = 1/343$. If we assume 16 bit precision, each integer has probability $2^{-16} = 1/65536$ - So 1/343 is a great improvement over the original data.

We can code the right halves of the original expressions as
    $R_1$ , $R_2$ , Add
    $R_1$ , $R_2$ , Mul
    $R_1$ , $R_2$ , Add

If there are many examples like these, it will be noted that the probability that the symbol in the first position is $R_1$, is close to one. Similarly, the probability that the second symbol is $R_2$ is close to one. This gives a probability of close to 1/7 for $R_1$, $R_2$, Add.

We can increase the probability of our code further by noting that in the expressions like
    35, 41, + : $R_1$, $R_2$, Add,
the last symbol is closely correlated with the third symbol - so that knowing the third symbol, we can assign very high probability to the final symbols that actually occur.

I have spoken of assigning high and low probabilities to various symbols. How does this relate to length of codes? If a symbol has probability p, we can devise a binary code for that symbol (a Huffman code) having approximately $-\log p$ bits in it. If we use many parallel codes for a symbol (as in our definition of algorithmic probability) we can have an equivalent code length of exactly $-\log p$.

The second example is a common kind of planning heuristic. When a problem is received by the system, it goes to the "planner" module. The module examines the problem and on the basis of this examination assigns 4 probabilities to it, $P_1$, $P_2$, $P_3$, and $P_4$.
$P_1$ is the probability that the quickest solution will be obtained by first breaking the problem into subproblems ("Divide and Conquer"), that must all be solved. Module $M_1$ breaks up the problem and sends the individual subproblems back to "planner".
$P_2$ is the probability that the quickest solution is obtainable by transforming the problem into several alternative equivalent problems. Module $M_2$ does these transformations and sends the resultant problems to "planner".
$P_3$ is the probability of solution by method $M_3$. $M_3$ could, for example, be a routine to solve algebraic equations.
$P_4$ is the probability of solution by method $M_4$. $M_4$ could, for example, be a routine to perform symbolic integration.

The operation of the System in assigning probabilities to various possible solution trials, looks much like the probability assignment process in a stochastic production grammar.

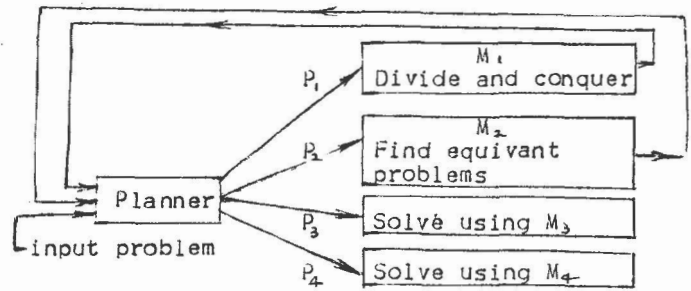

Because the outputs of $M_1$ and $M_2$ go back to "planner" we have a recursive system that can generate a search tree of infinite depth.

However, the longer solution trials have much less probability, and so, when we use the optimum search algorithm, we will tend to search the high probability trials first (unless they take too long to generate and test).

The third example is an anlysis of the concept "Analogy" from the viewpoint of Algorithmic probability.

The best known A.I. work in this area is that of T. Evans - a program to solve geometric analogy problems such as those used in intelligence tests [14]. We will use an example of this sort.

Given (a) △ ▽ , (b) ▽ △ (c) ⊥ ⊤
(d) ♂ ♀  (e) ♂ ♂ .

Is d or e more likely to be a member of the set a, b, c ?

We will devise short descriptions for the sets a, b, c, d and a, b, c, e, and show that a, b, c, d has a much shorter description.

The set a, b, c, d can be described by a single operator followed by a string of operands:
$Op_1$ = [print the operand, then invert operand and print it to the right].

A short description of a, b, c, d is then:
(1) $Op_1$ [description of △ , description of ▽, description of ⊥ , description of ♂ ].

To describe a, b, c, e we will also need the operator $Op_2$:
$Op_2$ = [print the operand, then print it again to the right].

A short description of a, b, c, e is then:
(2) $Op_1$ [description of △ ,
    description of ▽ , description of ⊥ ],
        $Op_2$ [description of ♂ ].

It is clear that (1) is a much shorter description than (2). We can make this analysis quantitative if we actually write out many descriptions of these two sets in some machine code.

If a, b, c, d is found to have descriptions of lengths 100, 103, 103, and 105, then its total probability will be
$2^{-103} \times (1 + 2^{-3} + 2^{-3} + 2^{-5}) = 1.28135 \times 2^{-100}$

If a, b, c, e has description lengths 105, 107, 108, it will have a total probability of
$2^{-100} \times [2^{-5} + 2^{-7} + 2^{-8}] = 2^{-100} \times .0429688$

The ratio of these probabilities is 29.8, so if these code lengths were correct, the symbols ♂ ♀ would have a probability 29.8 times as great as ♂ ♂ of being a member of the set a, b, c.

The concept of analogy is pervasive in many forms in science and mathematics. Mathematical analogies between mechanical and electrical systems make it possible to predict accurately the behavior of one by anlyzing the behavior of the other. In all of these systems, the pair of things that are analogous are obtainable by a common operator such as Op, operating on different operands. In all such cases, the kind of anlysis that was used in our example can be directly applied.

The fourth example is a discussion of clustering as an inductive technique.

Suppose we have a large number of objects, and the $i^{th}$ object is characterized by k discrete or continuous parameters, $(a_{i1}, a_{i2}, \ldots a_{ik})$. In clustering theory we ask, "is there some natural way to break this set of objects into a bunch of "clusters" (subsets), so that the elements within each cluster are relatively close to one another"?

Algorithmic probability theory regards this as a standard coding problem. The description of the set of objects consists, first, of a set of points in the space of parameters that corresponds to "centers of clusters". Each of the objects is then described by the name of its cluster and a description of how its parameters differ from that of its cluster center.

If the points of each cluster are distributed closely about their center, then we have achieved great compression of code by describing the cluster centers, followed by short codes giving the small distances of each point from its center.

The efficacy of this system depends critically on our formalism for describing the parameters. If we have any probabilistic information about the distribution of points this can be used to define good metrics to describe the deviations of points from their respective centers.

The fifth example is a description of frame theory as a variety of clustering.

Minsky's introduction to frames [15] treats them as a method of describing complex objects and storing them in memory. An example of a frame is a children's party. A children's party has many parameters that describe it: Who is giving the party? What time of day will it be given? Is it a birthday party? (i.e. Must we bring presents?) Will it be in a large room? What will we eat? What will we do? etc, etc. If we know nothing about it other than the fact that it's a party for children, each of the parameters will have a "default value" — this standard set of default parameters defines the "standard children's party". This standard can be regarded as a "center point" of a cluster space.

As we learn more about the party, we find out the true values of many of the parameters that had been given default assignments. This moves us away from the center of our cluster, with more complex (and hence less probable) descriptions of the parameters of the party. Certain of the parameters of the party can, in turn be described as frames, having sets of default values which may or may not change as we gain more information.

IV. PRESENT STATE OF DEVELOPMENT OF THE SYSTEM

How far have we gone toward realizing this system as a computer program?

Very little has been programmed. The only program specifically written for this system is one that compresses a text consisting of a long sequence of symbols. It first assigns probabilities to the symbols. Then new symbols are defined that represent short sequences appearing in the text [8, pp 232-240], and they are also assigned probabilities. The principles of this program are very useful, since in most bodies of data the main modes of compression are assignments of probabilities to symbols and defining new symbols.

We have studied compression of text by coding it as a stochastic context free grammar [8, pp. 240-253, also 10, pp. 276-277].

Some work has been done on devising training sequences as a means of inserting information into the probability distribution [9,12].

It has been possible to take an existing A.I. system and "retrofit" it with the present probabilistic approach [10, pp. 277-280]. Some of the best known work on mechanized induction is Winston's program for learning structures from examples [16]. It uses a training sequence of positive examples and close negative examples ("near misses"). After much computation, it is able to choose a (usually correct) structure corresponding to the examples.

The probabilistic form for this problem simplifies the solution considerably, so that probabilities for each possible structure can be obtained with very little calculation. The system is able to learn even if there are no negative examples — which is well beyond the capabilities of Winston's program. That probabilities are obtained rather than "best guesses" is an important improvement. This makes it possible to use the results to obtain optimum statistical decisions. "Best guesses" without probabilities are of only marginal value in statistical decision theory.

There are a few areas in which we haven't yet found very good ways to express information through probability distributions. Finding techniques to exand the expressive power of these distributions remains a direction of continued research.

However, the most important present task is to write programs demonstrating the problem solving capabilities of the system in the many areas where representation of knowledge in the form of probability distributions is well understood.

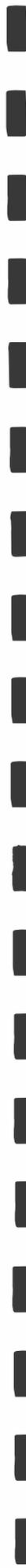